\pdfoutput=1

\documentclass[11pt]{article}

\usepackage{naacl2021}

\usepackage{times}
\usepackage{latexsym}
\usepackage{booktabs}
\usepackage{graphicx}
\usepackage{amsmath}
\usepackage{enumitem}
\usepackage{float}
\usepackage{multirow}
\usepackage{todonotes}
\usepackage{dingbat}

\usepackage[T1]{fontenc}

\usepackage[utf8]{inputenc}

\usepackage{microtype}

\title{Annotating and Modeling Fine-grained Factuality in Summarization}

\author{Tanya Goyal \and Greg Durrett \\
  Department of Computer Science \\
  The University of Texas at Austin \\
  {\tt tanyagoyal@utexas.edu, gdurrett@cs.utexas.edu}}

\begin{document}
\maketitle
\begin{abstract}
Recent pre-trained abstractive summarization systems have started to achieve credible performance, but a major barrier to their use in practice is their propensity to output summaries that are not faithful to the input and that contain factual errors. While a number of annotated datasets and statistical models for assessing factuality have been explored, there is no clear picture of what errors are most important to target or where current techniques are succeeding and failing. We explore both synthetic and human-labeled data sources for training models to identify factual errors in summarization, and study factuality at the word-, dependency-, and sentence-level. Our observations are threefold. First, exhibited factual errors differ significantly across datasets, and commonly-used training sets of simple synthetic errors do not reflect errors made on abstractive datasets like \textsc{XSum}. Second, human-labeled data with fine-grained annotations provides a more effective training signal than sentence-level annotations or synthetic data. Finally, we show that our best factuality detection model enables training of more factual \textsc{XSum} summarization models by allowing us to identify non-factual tokens in the training data.\footnote{Code and data available at \url{https://github.com/tagoyal/factuality-datasets}}
\end{abstract}
\section{Introduction}

Hallucination of unsupported or incorrect facts is a known shortcoming of current text generation and summarization models \cite{cao2018faithful, falke2019ranking}. This has been established for both abstractive summarization models \cite{maynez2020} and extractive summarization models \cite{kryscinski2020evaluating, falke2019ranking}. Past work has explored using off-the-shelf frameworks such as entailment models \cite{falke2019ranking} or QA systems \cite{durmus2020feqa, wang2020asking} to detect and sometimes correct errors in generated summaries. Another line of recent work has used synthetically generated data to specifically train models on the factuality detection task \cite{kryscinski2020evaluating, zhao2020reducing, goyal2020evaluating}. However, these efforts have focused on different datasets, summarization systems, and error types, often shedding little light on what errors state-of-the-art systems are actually making and how to fix them.

In this paper, we aim to answer two main questions. First, while synthetic data generation approaches are specifically designed for factuality evaluation, \textbf{do these align with actual errors made by generation models? We find the answer is no:} techniques using surface-level data corruption \cite{kryscinski2020evaluating, zhao2020reducing, cao2020factual} or paraphrasing \cite{goyal2020evaluating} target inherently different error distributions than those seen in actual model generations, and factuality models trained on these datasets perform poorly in practice. Furthermore, we show that different summarization domains, CNN/Daily Mail \cite{hermann2015teaching, nallapati2016abstractive} and XSum \cite{narayan2018don} (which differ in the style of summaries and degree of abstraction), exhibit substantially different error distributions in generated summaries, and the same dataset creation approach cannot be used across the board.

\begin{figure*}[h]
\centering
    \includegraphics[trim=75mm 77mm 40mm 55mm,scale=0.29,clip]{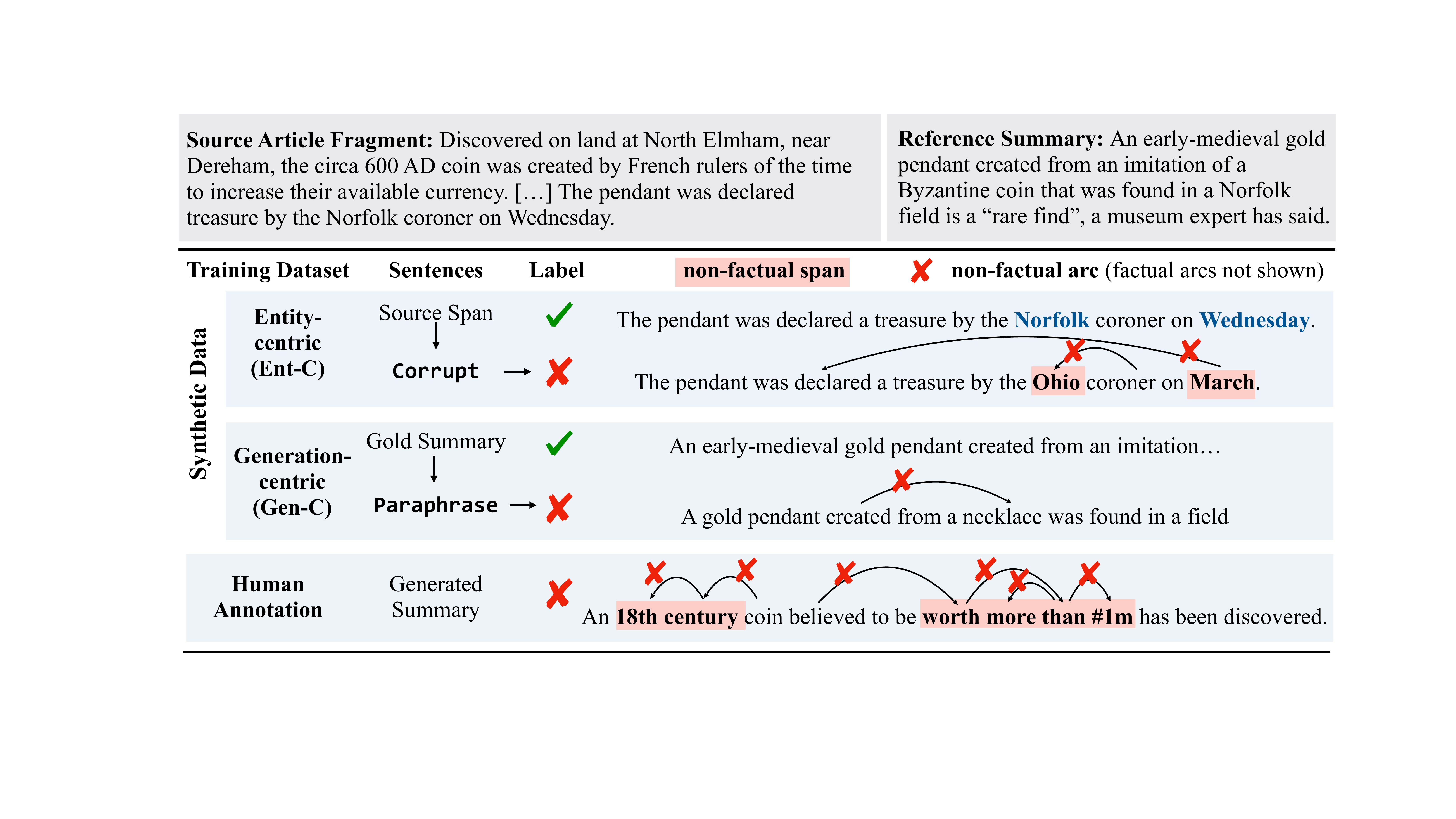}
    \caption{Examples from the synthetic and human annotated factuality datasets. The entity-centric and generation-centric approaches produce bad summaries from processes which can label their errors. All models can be adapted to give word-level, dependency-level, or sentence-level highlights, except for Gen-C.} 
    \label{fig:factuality-datasets}
\end{figure*}

Second, we investigate the best approach for modeling and learning factuality, particularly for highly abstractive summarization settings \cite{narayan2018don}. Specifically, we compare the utility of fine-grained human annotations (such as error highlighting at the word- or span-level) with sentence-level factuality annotations. We use a prior factuality detection model capable of leveraging such fine-grained annotations \cite{goyal2020evaluating} and show that these allow us to more reliably detect errors as well as localize those errors within generated texts. In fact, fine-grained human annotations are almost essential for any of our techniques to work well with high-performing summarizers in the challenging \textsc{XSum} setting.

Finally, we demonstrate a practical application for such error localization capabilities beyond interpretibility. Given noisy training data for summarization, we employ a modified training objective that leverages information about error spans in gold summaries, derived from factuality models, to train the summarizer. Our results show that models trained using this approach are inherently more factual than standard training objectives when dealing with error-prone gold datasets.

\section{Training Datasets to Compare}
\label{sec:training-data}
We first seek to answer how well synthetic training data can help address factuality errors observed in real summarization datasets. Figure~\ref{fig:factuality-datasets} shows a summary of the approaches we consider, which we describe in detail in Section \ref{sec:ent-c} and \ref{sec:gen-c}.

The summarization models we analyse are trained on two English-language domains: (1) \textbf{\textsc{XSum}}, an ``extreme'' summarization dataset from British Broadcasting Corporation (BBC) articles, where the first sentence of the article is treated as a summary of the article. These summaries are highly abstractive in nature: summarization models trained on this dataset have to learn to model long-range dependencies and may still be unable to recover all information in the gold summary. (2) \textbf{\textsc{Cnn/Dailymail}}, a multi-sentence abstractive summary dataset. The level of abstraction in this dataset is considerably lower and reference summaries exhibit high overlap with source articles \cite{zhang2018abstractiveness}. 

For both of these domains, we compare the distribution of factuality errors from \emph{synthetic training data} with the distribution of \emph{observed factuality errors } from models trained on that data. In Section~\ref{sec:model}, we further dive into factuality models' performance in these settings.

\subsection{Entity-centric Synthetic Data (Ent-C)}
\label{sec:ent-c}
A recent thread of work has focused on leveraging synthetic data transformations for evaluating factuality \cite{kryscinski2020evaluating}, imposing decoding-time constraints \cite{zhao2020reducing}, or post-correction of summaries \cite{cao2020factual}. Each of these approaches assumes that corruption strategies will yield useful non-factual summaries, while gold summaries are treated as factual. Figure~\ref{fig:factuality-datasets} illustrates this process: these approaches apply transformations to either the source article (shown) or a reference summary to obtain a corrupted summary (\emph{Ohio} instead of \emph{Norfolk}).

\begin{figure}[t]
\centering
    \includegraphics[trim=74mm 126mm 45mm 40mm,scale=0.32,clip]{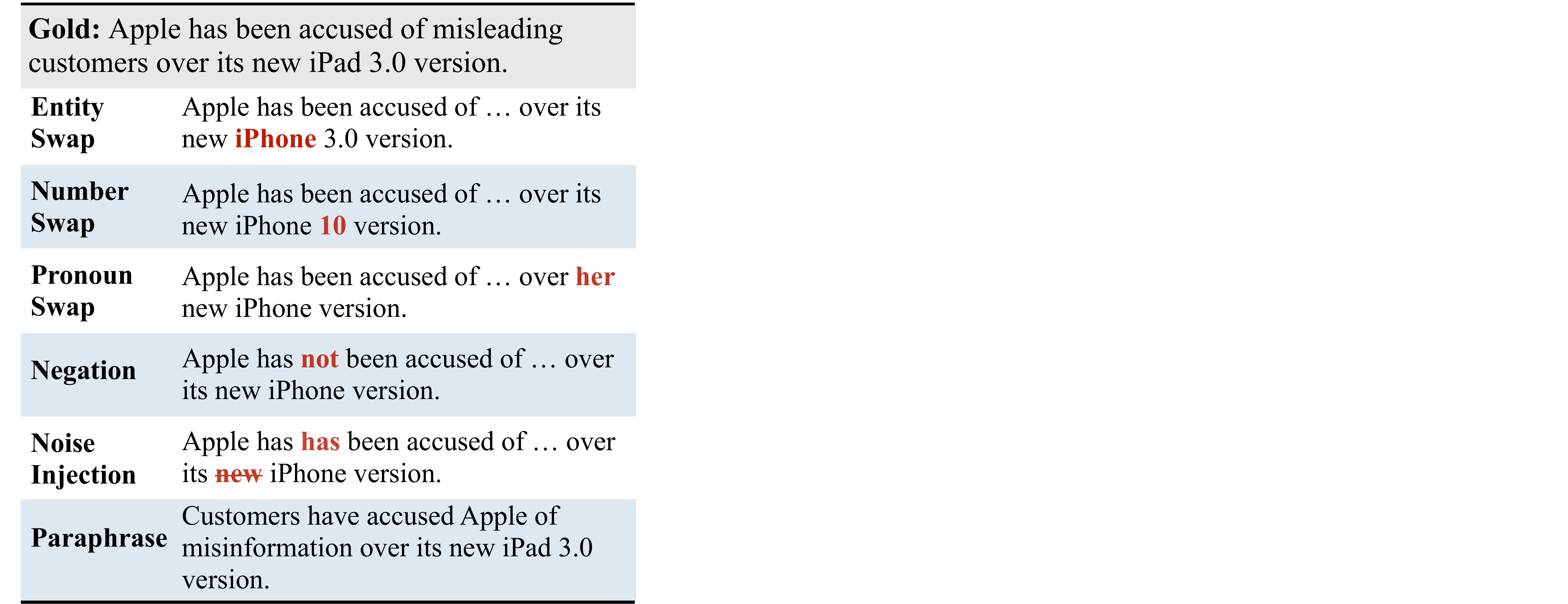}
    \caption{Set of transformations/data corruption techniques from \citet{kryscinski2020evaluating} used to generate training data for the entity-centric approach.}
    \label{fig:ent-c-examples}
\end{figure}

We call this set of approaches \textbf{entity-centric} because the transformations largely focus on perturbing entities and noun phrases and addressing these types of hallucinations. The approach from \citet{kryscinski2020evaluating} has the broadest set of transformations out of this line of prior work, so we follow them to generate training examples representative of this class of techniques. The data corruptions or transformations included are entity and number swapping, pronoun swapping, sentence negation, and arbitrary noise injection. Additionally, backtranslation is used to paraphrase summaries and further augment the dataset. Figure \ref{fig:ent-c-examples} illustrates the complete set of transformations applied to the reference summary to construct the synthetic dataset.

For \textsc{Cnn/Dm}, we use a dataset of 50k labeled pairs that is a subset of the data distributed by \newcite{kryscinski2020evaluating}; this subset is sufficient to reproduce the performance of their factuality classifier. We generate a similarly-sized dataset for \textsc{XSum}. Note that although the data creation procedure produces sentence-level annotations, since data corruptions are introduced in a rule-based manner, we can highlights spans within the summaries where the error was actually introduced to get span-level factuality annotations as well. Figure \ref{fig:factuality-datasets} illustrates these spans in red. The figure also demonstrates how to obtain dependency-level factuality judgements from error span highlights; what these mean and how these are derived is explained in Section \ref{sec:gen-c}.

\subsection{Generation-centric Synthetic Data (Gen-C)} 
\label{sec:gen-c}
\citet{goyal2020evaluating} introduce a different method for obtaining factuality annotations that more closely align with errors made by generation models. The core assumption of that \textbf{generation-centric} approach (see Figure~\ref{fig:factuality-datasets}) is that generated paraphrases at the bottom of a paraphrasing model's beam (the 10th-best paraphrase) are more likely to contain factual errors than 1-best generations, and new information in these generations can be labeled non-factual. Moreover, these generations align with realistic errors made by generation models, unlike purely synthetic entity swaps. In addition to sentence-level annotations, this approach also extracts \textbf{factuality labels corresponding to each dependency arc of the generated summary}. According to the definition given in \citet{goyal2020evaluating}, an arc is factual (or entailed) if the semantic relationship described by that particular dependency arc is entailed by the source article. Figure \ref{fig:factuality-datasets} shows a non-factual \emph{created $\rightarrow$ necklace} collapsed dependency arc. 

To adapt this data creation approach for our current experimental setting, we generated paraphrases of gold summaries using the paraphrase generation model of \citet{goyal-durrett-2020-neural}. We use the 10th-best generated summaries to generate both sentence-level and dependency-level annotations automatically. See Figure \ref{fig:factuality-datasets} for an example of this process. We generate 40k training examples for both \textsc{Cnn/Dm} and \textsc{Xsum} domains. 

\subsection{Types of supervision}
\label{subsec:deriving}
The two techniques, Ent-C and Gen-C, naturally generate annotations at different levels. We take steps to unify these formats to enable apples-to-apples comparison of them.

For Ent-C as well as human-labeled data (discussed later), we have access to span highlights within the summary that are non-factual with respect to the source article. From these, we can derive dependency-level annotations in the following way: for each arc in the summary, if either the head word or the child word is highlighted as non-factual, the dependency arc is annotated as non-factual. Otherwise, the arc is factual. This process is demonstrated in Figure \ref{fig:factuality-datasets}.

\begin{table}[t]
\small
\centering
\begin{tabular}{c|ccc}
\toprule
Dataset Source & Sent-level & Word-level & Dep-level \\ \midrule
Ent-C & \checkmark & \checkmark & \checkmark$^d$\\
Gen-C & \checkmark & & \checkmark\phantom{$^d$} \\
\textsc{Human-XSum} & \checkmark & \checkmark & \checkmark$^d$ \\
\bottomrule
\end{tabular}
\caption{Summary of the annotations available for each training dataset. \checkmark \ indicates that annotations at that granularity can be directly obtained from the data creation process. \checkmark$^d$ indicates that annotations can be derived.}
\label{table:dataset-annotations}
\end{table}

\begin{figure*}[h]
\centering
    \includegraphics[trim=70mm 70mm 45mm 40mm,scale=0.29,clip]{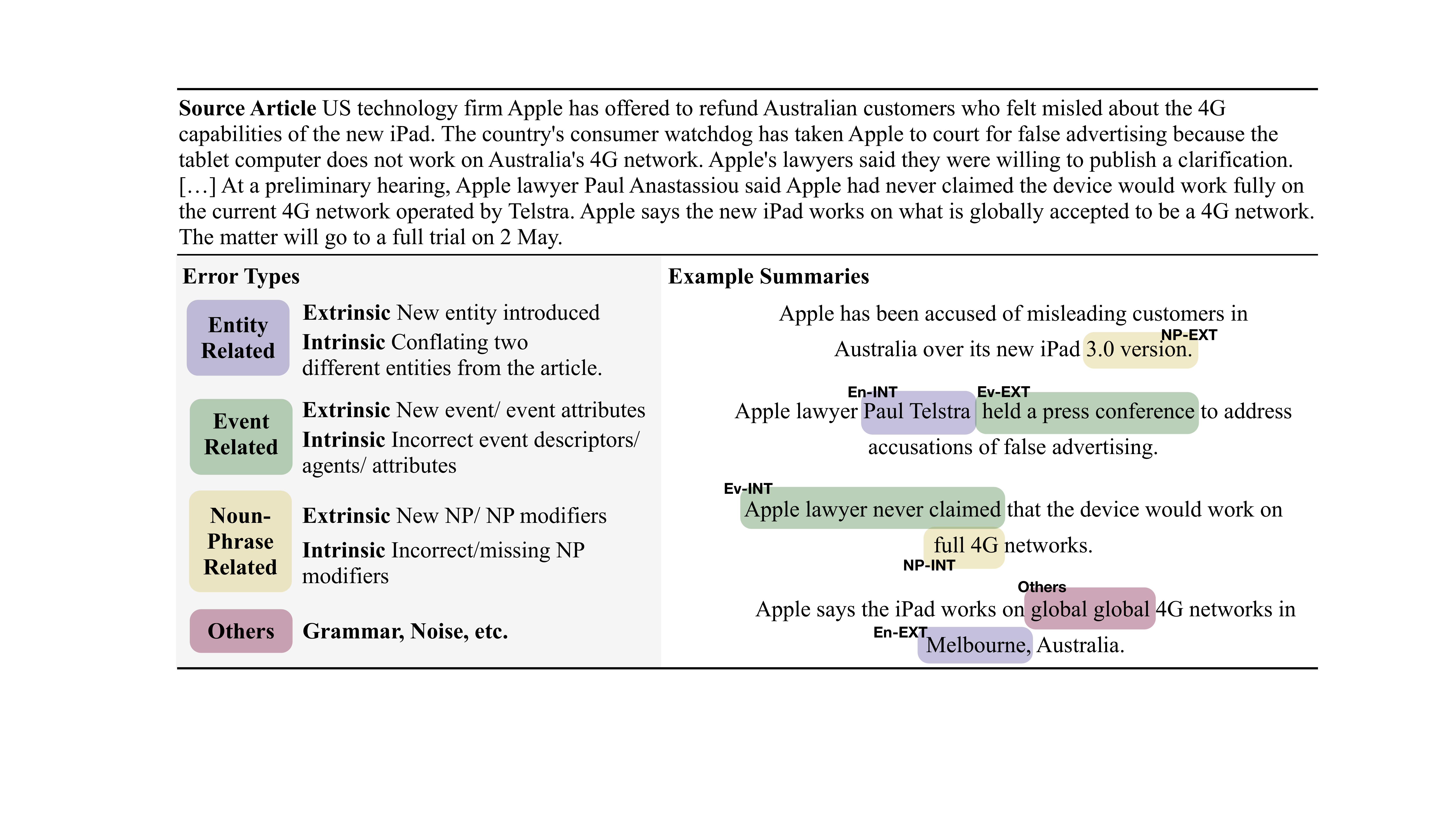}
    \caption{Taxonomy of error types considered in our manual annotation. On the right are example summaries with highlighted spans corresponding to the error types; the first summary is an actual \textsc{Bart} generated summary while others are manually constructed representative examples.}
    \label{fig:error-types}
\end{figure*}

Table \ref{table:dataset-annotations} gives a summary of the type of annotations available for the 3 types of training datasets. Mapping Gen-C dependency-level annotations to word-level classification decisions is less well-defined, so we do not attempt to do this. Our focus in this work will be on training sentence-level and dependency-level classification models, which is possible on all our sources of data.

\section{Analysis of Error Types}
\label{sec:error-discussion}
Past work using synthetic training data implicitly assumes that training a factuality model on such data will allow it to transfer to realistic settings.
We start by qualitatively analyzing the actual errors produced by summarization models to see how these align with the synthetic data, which helps us better understand this assumption.

We identify four broad categories of errors (see Figure \ref{fig:error-types}) that we will identify through manual inspection. Each of these categories is further divided into \textbf{Intrinsic} (errors that arise as a result of misinterpreting information from the source article) and \textbf{Extrinsic} (errors that hallucinate \emph{new} information or facts not present in the source article), following the characterization from \newcite{maynez2020}.
\begin{enumerate}[leftmargin=*]
    \item \textbf{Entity-Related}: errors specifically related to surface realization of named entities, quantities, dates, etc. Hallucination of new entities is an extrinsic error; incorrectly combining distinct entities from the source article is an intrinsic error (\textit{Paul Telstra} in Figure \ref{fig:error-types}). 
    \item \textbf{Event-Related}: errors with incorrect claims about events in the summary, such as predicates with arguments filled by incorrect entities.  Hallucinations of new events (\textit{held a press conference} in Figure \ref{fig:error-types}) are extrinsic; mixed-up attributes from within the source article are intrinsic (\textit{apple lawyer never claimed} in Figure \ref{fig:error-types}, incorrect agent).
    \item \textbf{Noun Phrase-Related}: errors related to noun phrases other than the entity-specific errors. Examples include hallucinating new NP modifiers (extrinsic) or combining with a wrong modifier from the article (intrinsic).
    \item \textbf{Other Errors}: errors such as ungrammatical text, repeated words, highly erroneous spans, etc. that don't fall into one of the above categories. These are not broken down by intrinsic/extrinsic.
\end{enumerate}

\begin{figure}[t]
\centering
    \includegraphics[trim=87mm 25mm 300mm 25mm,scale=0.27,clip]{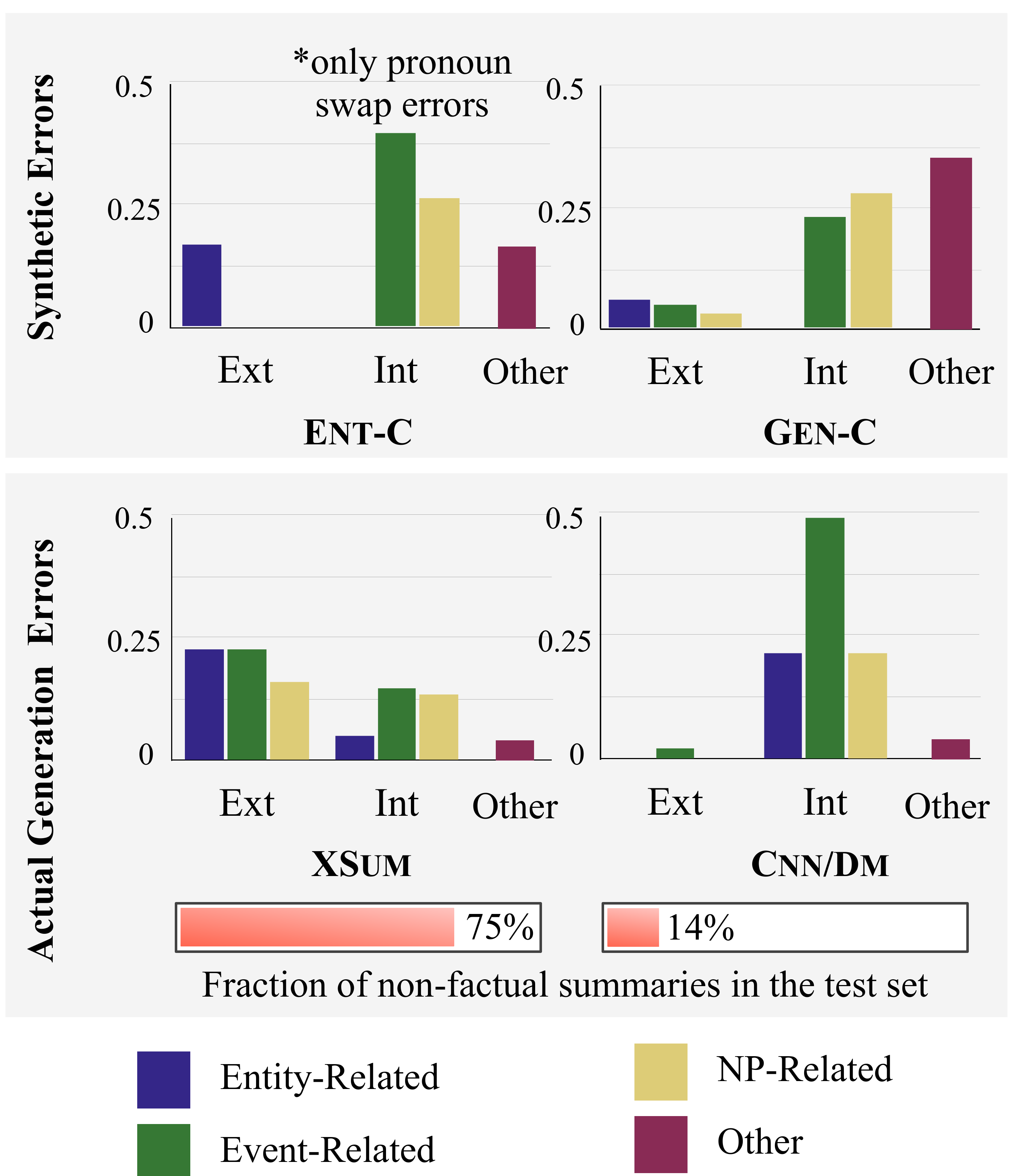}
    \caption{Fractions of examples in each dataset exhibiting different error types (note a single example may have multiple errors).
    The graphs show a significant mismatch between the error distributions of actual generation models and synthetic data corruptions.}
     \label{figure:error-distributions}
\end{figure}

Our taxonomy of summarization errors differs from that of \citet{lux-etal-2020-truth}: theirs is targeted at the effects on the reader, whereas ours is more directly tied to the grammatical role of the error, which we believe is more useful to improve our data and our systems. We use the above taxonomy to annotate examples from both summarization domains. For \textsc{XSum}, we use the state-of-the-art \textsc{Bart} model \cite{lewis-etal-2020-bart} to generate summaries followed by manual annotation (100 examples). For \textsc{Cnn/Dm}, annotation was done on the 50 summaries across 10 different models collected by \citet{kryscinski2020evaluating}. We additionally do this annotation for the artificially introduced errors in Ent-C and Gen-C.\footnote{Discussion of inter-annotator agreement is included in Appendix \ref{app:annotation}.}

\paragraph{Results} Figure \ref{figure:error-distributions} shows the distribution of errors for these different settings. First, we see that \textbf{summarization models from different domains make substantially different types of errors}. Models trained on \textsc{XSum} learn to hallucinate new content and consequently produce extrinsic errors: 60\% of the errors made by \textsc{Bart} models are extrinsic. One reason for this is that the \textsc{XSum} data was automatically constructed and contains gold summaries that are noisy or non-factual (75\% of gold summaries, according to \citet{maynez2020}). In addition to this, the gold summaries are also highly abstractive, and \textbf{XSum-trained summarization models learn to combine information from different parts of an article}, leading to models making long-range dependency errors. This misinterpretation of content is largely responsible for the 40\% of the errors which are intrinsic.

On the other hand, the \textsc{Cnn/Dm} summarization datasets contain human written gold summaries and are therefore generally much more reliable. The models trained on this dataset reflects that. Only 14\% of the generated summaries contains errors in the \textsc{Cnn/Dm} validation set from  \cite{kryscinski2020evaluating}. Of these 14\%, \textbf{the bulk of the errors produced are intrinsic errors}, primarily event-related caused by sentence compression or fusion, which is common in this dataset \cite{lebanoff-etal-2019-analyzing}. For example, \emph{the two Delaware boys are in critical condition at the U.S.~Virgin Islands} should instead be \emph{...at the hospital after a trip to the U.S.~Virgin Islands.} The generation models rarely makes extrinsic hallucinations, and we observed that these are even less common in recent systems like PEGASUS \cite{pegasus2020}. This aligns with the findings from prior work analysing summarization models \cite{fabbri2020summeval}.

Comparing these with synthetic error distributions, we can see that \textbf{synthetic datasets do not reflect the error distributions of actual generation models}. To the extent that Ent-C covers intrinsic event-related errors, these are almost exclusively from pronoun swaps. 
Moreover, because \textsc{Cnn/Dm} and \textsc{XSum} feature such different errors, a synthetic dataset inspired by observed errors on one setting is not likely to be effective on the other.
Later (in Section \ref{sec:exp-synthetic-datasets}), we provide further evidence of this mismatch for both datasets: models trained on this synthetic data perform poorly when evaluated on actual generation errors. Also, models trained on human annotated \textsc{XSum} training data do not transfer to the \textsc{Cnn/Dm} domain.

\section{Factuality Models to Compare}
\label{sec:model}
Next, we investigate how factuality models trained on these synthetic datasets perform on real generation errors. 
Given a document $D$, a factuality model predicts whether all the information in a generated summary $S$ is supported by the source document $D$.\footnote{Factuality is ill-defined: whether inferences, world knowledge, implicatures, etc. are viewed as factual is not standardized and is dependent on human annotators for each dataset or task. However, existing generation models only rarely exhibit tricky cases along these dimensions.} We consider two factuality modeling formulations: (1) a \textbf{Sentence-Factuality} model that makes a factuality judgment at the entire summary-level, and (2) an \textbf{Arc-Factuality} model \cite{goyal2020evaluating} that makes independent factuality judgments for dependency arcs of the generated summary, which are then combined to obtain a sentence-level decision. This helps in localizing factuality errors and was shown to be more effective than sentence-level models in prior work.\footnote{We describe models for single-sentence summaries to align with the available human-annotated test set (described later in Section \ref{sec:exp-synthetic-datasets}). However, it is straightforward to extend these frameworks for multi-sentence summaries.}

\subsection{Sentence-Factuality Model} 
\label{sec:sent-factuality-model}
Prior work \cite{kryscinski2020evaluating} used a \textsc{Bert}-based sequence-pair classification model \cite{devlin2019bert} as follows: the source document $D$ and the generated summary $S$ are concatenated and fed into a pre-trained transformer encoder model (\textsc{Bert, Electra}, etc.). The representation of the [CLS] token is fed into a linear and softmax layer that outputs a probability distribution over the output labels ($y=\{\textrm{Factual, Non-Factual}\}$). This model can be trained on any data with summary-level factuality labels.

\begin{figure}[t]
\includegraphics[trim=60mm 177mm 0mm 95mm,scale=0.40,clip]{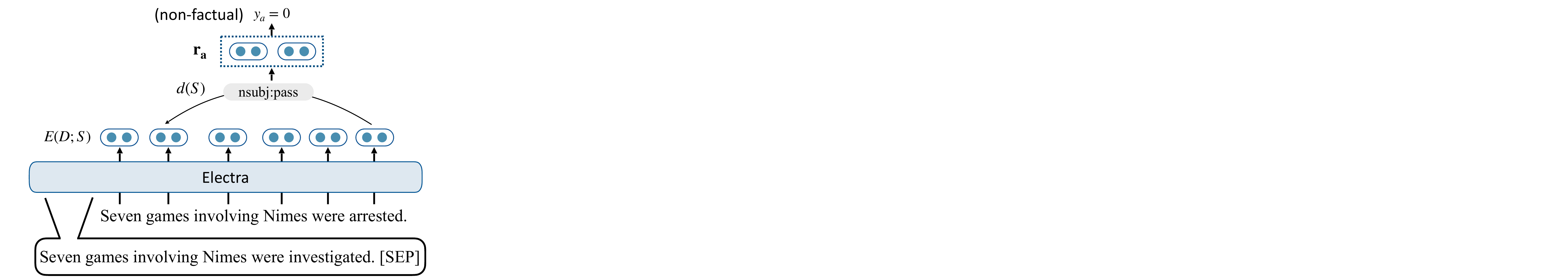}
\centering
\caption{The dependency arc entailment (DAE) model from \cite{goyal2020evaluating}. A pre-trained encoder is used to obtain arc representations; these are used to predict arc-level factuality decisions.}
\label{fig:dae_model}
\end{figure}

\subsection{Arc-Factuality model}
\label{sec:dae-model}
The \textbf{Dependency Arc Entailment (DAE)} model \cite{goyal2020evaluating} evaluates factuality at the dependency arc level. Let $d(S)$ be the dependency-parse of the generated summary $S$. For each arc $a \in d(S)$, the DAE model predicts whether the relationship described by the arc is entailed by the input document. Note that these factuality judgements are made \emph{independently} for each arc in the summary, and can differ across arcs within the same summary. For instance, in the example in Figure \ref{fig:dae_model}, the arc $arrested \leftarrow games$ is non-factual: in context, it is not the case that the games are being arrested. However, the arc $seven \leftarrow games$ is supported by the input (there are seven games) and hence, entailed. 

The model architecture is detailed in Figure \ref{fig:dae_model}. First, the document $D$ and summary $S$ are concatenated and fed through a pre-trained encoder $E$. Arc representations $\mathbf{r}_a$ are derived for each dependency arc $a \in d(S)$: $\mathbf{r}_a =  [E(D;S)_{a_h};E(D;S)_{a_c}]$. Here, ${a_h}$ and ${a_c}$ correspond to the head and child words of arc $a$ respectively. The arc representation $\mathbf{r}_a$ is fed into a classification layer that outputs a probability distribution over the output labels ($y_{a}=\{\textrm{Factual, Non-Factual}\}$). Finally, \textbf{summary-level} judgments are extracted from these arc-level decisions: if \emph{any} dependency arc is non-factual, the generated summary is labeled as non-factual.

The DAE model is trained from arc-labeled examples of the form $(D, S, \{y_a \}_{a\ \in\ d(S)})$. These are derived from either synthetic or human-labeled data, as described in Section \ref{sec:training-data}.

\paragraph{DAE with weak supervision (DAE-Weak)} DAE training requires gold annotations at the dependency-level; however, such fine-grained annotations may not always be available. We extend the DAE framework to address this. The core idea behind our approach is that the sentence-level labels naturally impose loose constraints on the arc-level labels.

The constraints are as follows: for a factual example, \emph{all} individual arcs in the summary must be factual. For a non-factual example, \emph{at least one arc} must be non-factual, and this arc should be one \emph{not} present in the source document. The DAE-Weak model is trained to maximize the marginal likelihood of all labelings that obey these constraints.

Let $F$ be the set of all arcs that should be factual (contains all arcs with sent-label = 1 and arcs common with the source article for sent-label = 0). The above constraints are formulated as the following training objective:
\begin{align*}
    \mathcal{L} =& \log \left[\prod_{a \in F} P(y_a = 1 \mid D, S) \right]\\ &+ \log \left[1 - \prod_{a \in D(S) \backslash F} P(y_a = 1\mid D, S)\right]
\end{align*}
The second term in the above equation is the probability of predicting at least one non-factual arc in a non-factual summary.\footnote{This techniques resembles posterior regularization \cite{ganchev2010posterior}; however, these constraints are enforced in a hard way on individual examples rather than in expectation at the corpus level. It can also be viewed as an instance of constraint-driven learning \cite{chang2007guiding}.}

\section{Experiments}

\subsection{Evaluation of Synthetic Training Datasets}
\label{sec:exp-synthetic-datasets}
\paragraph{\textsc{Cnn/Dailymail}}
First, we compare the performance of the three models (Sent-Factuality, DAE and DAE-Weak) trained on the two synthetic factuality datasets (outlined in Section \ref{sec:training-data}) on the \textsc{Cnn/Dailymail} domain. We compare their performance on the human-annotated test dataset from \citet{kryscinski2020evaluating}. The test set  contains human-annotated sentence-level factuality judgements for 503 (article, summary) pairs for summaries generated using 10 different generation models. We use the validation set provided by the authors to choose the best model checkpoint across all settings. Similar to the original paper, we report \textbf{class-balanced accuracy values}.

\begin{table}[t]
\small
\centering
\begin{tabular}{r|c|c}
\toprule
 & Ent-C & Gen-C \\  \midrule
Majority Label & 50 & 50 \\ \midrule
\citet{kryscinski2020evaluating} & 74.1 & - \\
Sent-Factuality & 72.3  & 64.4 \\
DAE &  \textbf{76.7} & \textbf{72.1} \\
DAE-Weak & 75.2 & 71.1 \\
\bottomrule
\end{tabular}
\caption{Label-balanced accuracy of factuality models when trained on synthetic factuality training datasets in the \textsc{Cnn/Dailymail} domain. Performance is reported on the human-annotated test set from \citet{kryscinski2020evaluating}.} 
\label{table:cnn-synthetic}
\end{table}

Table \ref{table:cnn-synthetic} outlines our results. The results show that models trained on Ent-C perform slightly better than those trained on Gen-C, but many of the systems are in the same range, with accuracy values of around 75\%. However, the reported accuracy values on held-out Ent-C/Gen-C examples are consistently over 90\% (results included in Appendix \ref{app:cnndm-held-out}). This demonstrates that while models trained on these factuality datasets are able to fit the synthetic data distributions well, these are inherently different from actual generation errors. The Appendix also includes graphs of how the human annotated dev set performance varies with training iterations, showing that constant performance on the held-out training set corresponds with highly fluctuating performance on the human annotated data, further indicating that these settings are not identical.

\paragraph{\textsc{XSum}} Next, we similarly evaluate the synthetic datasets and factuality models on the more challenging \textsc{XSum} domain. Again, we evaluate on a human annotated dataset collected by prior work \cite{maynez2020}. The dataset contains span highlights indicating hallucinated/incorrect content or information with respect to the source article for 4 different summarization models trained on the \textsc{XSum} domain (as well as for gold summaries). Figure \ref{fig:factuality-datasets} illustrates this. Similar to prior work, if any word in a summary is marked as hallucinated, we mark the sentence as non-factual. Therefore, for \textsc{XSum-Human}, the annotation is available at both the sentence-level and span-level.

In total, this dataset contains 2500 $(A, S)$ pairs (along with their factuality labels). We use 500 examples from these to construct our test dataset. The remaining 2000 examples are used to train models, explained in Section \ref{sec:xsum-human}.

\begin{table}[t]
\small
\centering
\begin{tabular}{r|c@{\hskip 0.1in}cc@{\hskip 0.1in}c}
\toprule
Train Data & Majority & Sent-Fact & DAE & DAE-Weak \\ \midrule
Ent-C & 50 & 50.9 & 51.2 & 53.6 \\
Gen-C & 50 & 54.2 & 53.0 & 51.6\\
\bottomrule
\end{tabular}
\caption{Performance of factuality models trained on synthetic factuality datasets in the \textsc{XSum} domain. Label-balanced accuracy is reported on 500 examples from the human-annotated test set from \citet{maynez2020}.}
\label{table:xsum-synthetic}
\end{table}

Table \ref{table:xsum-synthetic} outlines the results. Unlike on \textsc{Cnn/Dm}, we see that all models trained on synthetic factuality datasets perform very poorly, achieving close to the majority label baseline. Again, the performance on the held-out synthetic datasets was observed to be very high (see Appendix \ref{app:cnndm-held-out}). \textbf{There is a fundamental difference between the errors that are produced by \textsc{XSum} summarization models and those introduced by artificial data corruption mechanisms.} Other data that more closely resembles the generation errors is needed to train factuality models in this setting.

\subsection{Human Annotated Dataset Evaluation}
\label{sec:xsum-human}
To investigate whether human annotated data is useful to train factuality models, we train our 3 factuality models on the remaining 2000 human annotated examples from \textsc{XSum-Human}. In order to train DAE model on this dataset, we use the span highlights to derive dependency-level gold annotations, using the same strategy from \ref{subsec:deriving} (illustrated in Figure \ref{fig:factuality-datasets}).

\begin{table}[t]
\small
\centering
\begin{tabular}{r|c}
\toprule
Model & Balanced-Acc \\ \midrule
Sent-Factuality & 65.6 \\
DAE &  \textbf{78.7} \\
DAE-Weak & 70.9 \\ 
\bottomrule
\end{tabular}
\caption{Comparison of different factuality models when trained on human annotated data and evaluated on \textsc{XSum} (compare to Table~\ref{table:xsum-synthetic}). Fine-grained annotations provide a big boost in performance.}
\label{table:xsum-human}
\end{table}

The results are shown in Table \ref{table:xsum-human}. Comparing these with results from Table \ref{table:xsum-synthetic}, we see that a small number of human annotated examples can outperform large auto-generated training datasets by a large margin. Notably, we see that \textbf{availability of fine-grained factuality annotations significantly boosts performance}, with models that leverage that information (DAE) significantly outperforming sentence-level models. Even in the absence of fine-grained annotations, we see that the DAE-Weak model that decomposes the error computation and explicitly tries to localize errors is better than the sentence-level model.

However, \textbf{these factuality models do not transfer to \textsc{Cnn/Dm}}: the best model achieves an accuracy of 55.9, substantially lower than 76.7\% in Table~\ref{table:cnn-synthetic}. This demonstrates that summarization models make different types of errors on different domains, and data collection and modelling efforts for factuality should account for these differences. 

\section{Localization of errors}
Our evaluation so far has focused on the sentence-level performance of factuality models. Next, we evaluate the models' ability to localize errors within the generated summary as well as show how such a capability can be leveraged to train less error-prone summarization models. 
\subsection{Localizing Factuality on \textsc{XSum}}
We evaluate the error localization performance of the models at two granularity levels:
(1) \textbf{Dependency arc-level} and (2) \textbf{Word-level}.\footnote{We can approximately extract word-level decision from the dependency-level predictions: if any arc containing word $w$ is non-factual, then $w$ is non-factual; otherwise, it is factual.} Table \ref{table:dae-xsum-dep-level} outlines the results of our experiments. 

\begin{table}[h]
\small
\centering
\begin{tabular}{r|ccc}
\toprule
Model & Precision & Recall & F1 \\ \midrule
\multicolumn{4}{c}{\textbf{Dependency-level}} \\ \midrule
DAE & 69.7 & 78.2 & 73.7 \\
DAE-Weak & 54.9 & 76.6 & 63.9\\ \midrule
\multicolumn{4}{c}{\textbf{Word-level}} \\ \midrule
DAE & 57.5 & 74.7 & 65.0 \\
DAE-Weak & 56.2 & 62.3 & 59.1 \\ 
DAE (best-ckpt) & 62.0 & 83.9 & 71.3 \\
\bottomrule
\end{tabular}
\caption{Error localization comparison of the different factuality models. The DAE model achieves high recall for both word-level and dependency-level factuality.} 
\label{table:dae-xsum-dep-level}
\end{table}

The DAE model outperforms the DAE-Weak model at both levels of granularity. This reiterates our earlier claim that \textbf{fine-grained annotations lead to better factuality models with more reliable localization}. However, the DAE-Weak model is able to achieve comparable recall at the dependency-level; both models are more recall-oriented, which is desirable for certain applications.

For Section~\ref{sec:downstream}, we select our DAE model's best checkpoint on the test data (\textbf{best-ckpt}), which achieves a recall of 83.9, a significant gain if we directly optimize for this metric.

\subsection{Downstream Applications}
\label{sec:downstream}
Localizing errors potentially allows for post-hoc correction \cite{zhao2020reducing,cao2020factual}; however, repairing a summary to be fully factual is a very hard problem and past work has focused on a subset of errors as a result.
Instead, we show that even our imperfect error localization techniques can be used to meaningfully improve the \emph{training} data for summarization. 
We use our DAE model to identify unsupported facts in the \textsc{XSum} training data and ignore the corresponding tokens when training our summarization model.

\paragraph{Training on a subset of tokens} Summarization models are trained to maximize the log likelihood of the summary given the source article: $\mathcal{L} = \sum_{i=1:|S|} \log p(S_i|D,S_{1:i-1})$. When a word in the summary is non-factual, training on it encourages the model to hallucinate new content. In our approach, we modify the training objective to only maximize the likelihood of \emph{factual} words in the summary, factuality being determined by the DAE model from  the previous sections: $\mathcal{L} = \sum_{i=1:|S|} M_{i} \log p(S_i|D,S_{1:i-1})$ where $M_i = 1$ if the word $w_i$ is factual, otherwise $M_i = 0$. A similar objective has been used by prior work \cite{song2020controlling} to encourage the model to copy words present in the source.

We compare our approach with two systems: a baseline model trained without this masking and a model using the loss truncation technique well-suited for noisy datasets from \citet{kang-hashimoto-2020-improved}. All models are trained on 50k examples using \textsc{Bart} summarization model initialized from the \textsc{Bart-Xsum-Large} checkpoint. For all these approaches, summaries generated on the original \textsc{XSum} test set (11k examples) are compared.\footnote{To ensure fair comparison between the different models, we removed the examples from \textsc{XSum-Human} used to train the factuality models from our test set.}

\paragraph{Evaluation} First, we use our trained DAE model to evaluate the performance of our summarization models. That is, we generate summaries for all examples in the test set using the three models; the DAE model is then used to compute the word error rate (fraction of words determined to be non-factual according to the DAE model) and the sentence error rate (fraction of sentences determined to be non-factual). Table \ref{table:error-rates} outlines the results, which show that our DAE-masked training leads to better factuality performance.

\begin{table}[t]
\small
\centering
\begin{tabular}{r|c@{\hskip 0.1in}c@{\hskip 0.1in}c}
\toprule
Model & Word-ER $\downarrow$ & Sent-ER $\downarrow$ & Human $\uparrow$ \\ \midrule
Baseline & 32.2 & 74.0 & 37.4\\
Loss trunc  & 31.1 & 70.9 & 39.1\\
DAE-based (ours) & \textbf{23.7} & \textbf{61.4} & \textbf{46.5}\\ 
\bottomrule
\end{tabular}
\caption{Comparison of the different summarization models. Our proposed approach achieves significantly lower word error rates, sentence error rates and are rated higher by human annotators. }
\label{table:error-rates}
\end{table}

Next, we perform human evaluation to compare the factuality of summaries generated by the three models using Amazon Mechanical Turk. We randomly sampled 50 articles from the test set and generated summaries corresponding to the 3 models.\footnote{See Appendix \ref{app:mtruk} for more details about the task design.}. We asked 7 human annotators to classify each (article, summary) pair as either factual (score = 1) or non-factual (score = 0). An average score is computed for each summary by aggregating the 7 annotator scores. Table \ref{table:error-rates} reports the average summary scores for the 50 (article, summary) pairs across the 3 summarization models. The results show that the proposed approach outperforms both the baseline model and the loss truncation approach. This demonstrates that \textbf{factuality models trained on a small number of annotated examples can be used to train \emph{factual} summarization models, even when the underlying summarization dataset is noisy}. 

\section{Related Work}
Earlier work on abstraction \cite{barzilay-etal-1999-information,carenini-cheung-2008-extractive} and compression \cite{knight-marcu-2000-sentence-compression,berg-kirkpatrick-2011-jointly,woodsend-lapata-2012-multi-aspect,durrett-etal-2016-learning} in summarization has typically focused evaluation on content selection and grammaticality, with little heed paid to factuality. Human evaluation similarly focused on content selection \cite{gillick-liu-2010-non}. Methods such as Pyramid \cite{nenkova-passonneau-2004-evaluating} that could have in principle been used to evaluate factuality were primarily used to understand content selection. 

Recent work has explored different methods for enforcing factuality: modifying the model, such as encoding SRL structures in the input \cite{cao2018faithful}, post-hoc correction \cite{dong2020multi}, or constrained decoding \cite{song2020joint,mao2020constrained}. However, these techniques fundamentally struggle to handle the whole range of factual errors; factuality is a fuzzy notion and cannot be easily encapsulated into a set of discrete rules.

Faithfulness and factuality have also been tackled in related tasks, including summarizing radiology reports \cite{zhang-etal-2020-optimizing} and data-to-text generation tasks \cite{tian2019sticking}. Another recent line of work has looked at fact verification \cite{thorne-etal-2018-fever, nie2019combining, atanasova-etal-2020-generating-fact}. In this literature, the claims are usually human-authored and a straightforward statement of a fact, whereas generated summaries might feature claims buried in nominal modifiers like \emph{two-time winner}.

\section{Conclusion}
In this work, we showed that existing synthetic datasets are not well-suited to factuality evaluation of recent summarization models (like \textsc{Bart}) in challenging domain (like \textsc{XSum}). Models trained on human-annotated data, especially those that leverage fine-grained annotations, can enable training of more factual summarization models. We hope future work will explore better modeling and data creation to address the pressing issues in current systems.

\section*{Acknowledgments}
This work was partially supported by NSF Grant IIS-1814522, a gift from Salesforce Inc, and an equipment grant from NVIDIA. Thanks as well to Jiacheng Xu and the anonymous reviewers for their helpful comments.

\bibliography{anthology,custom}

\begin{thebibliography}{41}
\expandafter\ifx\csname natexlab\endcsname\relax\def\natexlab#1{#1}\fi

\bibitem[{Atanasova et~al.(2020)Atanasova, Simonsen, Lioma, and
  Augenstein}]{atanasova-etal-2020-generating-fact}
Pepa Atanasova, Jakob~Grue Simonsen, Christina Lioma, and Isabelle Augenstein.
  2020.
\newblock Generating fact checking explanations.
\newblock In \emph{Proceedings of the 58th Annual Meeting of the Association
  for Computational Linguistics}, pages 7352--7364.

\bibitem[{Barzilay et~al.(1999)Barzilay, McKeown, and
  Elhadad}]{barzilay-etal-1999-information}
Regina Barzilay, Kathleen~R. McKeown, and Michael Elhadad. 1999.
\newblock \href {https://doi.org/10.3115/1034678.1034760} {Information fusion
  in the context of multi-document summarization}.
\newblock In \emph{Proceedings of the 37th Annual Meeting of the Association
  for Computational Linguistics}, pages 550--557, College Park, Maryland, USA.
  Association for Computational Linguistics.

\bibitem[{Berg-Kirkpatrick et~al.(2011)Berg-Kirkpatrick, Gillick, and
  Klein}]{berg-kirkpatrick-2011-jointly}
Taylor Berg-Kirkpatrick, Dan Gillick, and Dan Klein. 2011.
\newblock {Jointly Learning to Extract and Compress}.
\newblock In \emph{Proceedings of the Annual Meeting of the Association for
  Computational Linguistics (ACL)}, pages 481--490.

\bibitem[{Cao et~al.(2020)Cao, Dong, Wu, and Cheung}]{cao2020factual}
Meng Cao, Yue Dong, Jiapeng Wu, and Jackie Chi~Kit Cheung. 2020.
\newblock Factual error correction for abstractive summarization models.
\newblock In \emph{Proceedings of the 2020 Conference on Empirical Methods in
  Natural Language Processing (EMNLP)}, pages 6251--6258.

\bibitem[{Cao et~al.(2018)Cao, Wei, Li, and Li}]{cao2018faithful}
Ziqiang Cao, Furu Wei, Wenjie Li, and Sujian Li. 2018.
\newblock Faithful to the original: Fact aware neural abstractive
  summarization.
\newblock In \emph{Proceedings of the AAAI Conference on Artificial
  Intelligence}, volume~32.

\bibitem[{Carenini and Cheung(2008)}]{carenini-cheung-2008-extractive}
Giuseppe Carenini and Jackie C.~K. Cheung. 2008.
\newblock \href {https://www.aclweb.org/anthology/W08-1106} {Extractive vs.
  {NLG}-based abstractive summarization of evaluative text: The effect of
  corpus controversiality}.
\newblock In \emph{Proceedings of the Fifth International Natural Language
  Generation Conference}, pages 33--41, Salt Fork, Ohio, USA. Association for
  Computational Linguistics.

\bibitem[{Chang et~al.(2007)Chang, Ratinov, and Roth}]{chang2007guiding}
Ming-Wei Chang, Lev Ratinov, and Dan Roth. 2007.
\newblock Guiding semi-supervision with constraint-driven learning.
\newblock In \emph{Proceedings of the 45th annual meeting of the association of
  computational linguistics}, pages 280--287.

\bibitem[{Devlin et~al.(2019)Devlin, Chang, Lee, and
  Toutanova}]{devlin2019bert}
Jacob Devlin, Ming-Wei Chang, Kenton Lee, and Kristina Toutanova. 2019.
\newblock Bert: Pre-training of deep bidirectional transformers for language
  understanding.
\newblock In \emph{Proceedings of the 2019 Conference of the North American
  Chapter of the Association for Computational Linguistics: Human Language
  Technologies}, pages 4171--4186.

\bibitem[{Dong et~al.(2020)Dong, Wang, Gan, Cheng, Cheung, and
  Liu}]{dong2020multi}
Yue Dong, Shuohang Wang, Zhe Gan, Yu~Cheng, Jackie Chi~Kit Cheung, and Jingjing
  Liu. 2020.
\newblock Multi-fact correction in abstractive text summarization.
\newblock In \emph{Proceedings of the 2020 Conference on Empirical Methods in
  Natural Language Processing (EMNLP)}, pages 9320--9331.

\bibitem[{Durmus et~al.(2020)Durmus, He, and Diab}]{durmus2020feqa}
Esin Durmus, He~He, and Mona Diab. 2020.
\newblock {FEQA: A Question Answering Evaluation Framework for Faithfulness
  Assessment in Abstractive Summarization}.
\newblock In \emph{Proceedings of the 58th Annual Meeting of the Association
  for Computational Linguistics}, pages 5055--5070.

\bibitem[{Durrett et~al.(2016)Durrett, Berg-Kirkpatrick, and
  Klein}]{durrett-etal-2016-learning}
Greg Durrett, Taylor Berg-Kirkpatrick, and Dan Klein. 2016.
\newblock {Learning-Based Single-Document Summarization with Compression and
  Anaphoricity Constraints}.
\newblock In \emph{Proceedings of the Annual Meeting of the Association for
  Computational Linguistics (ACL)}, pages 1998--2008.

\bibitem[{Fabbri et~al.(2021)Fabbri, Kry{\'s}ci{\'n}ski, McCann, Xiong, Socher,
  and Radev}]{fabbri2020summeval}
Alexander~R Fabbri, Wojciech Kry{\'s}ci{\'n}ski, Bryan McCann, Caiming Xiong,
  Richard Socher, and Dragomir Radev. 2021.
\newblock {SummEval: Re-evaluating summarization evaluation}.
\newblock \emph{Transactions of the Association for Computational Linguistics
  (TACL)}.

\bibitem[{Falke et~al.(2019)Falke, Ribeiro, Utama, Dagan, and
  Gurevych}]{falke2019ranking}
Tobias Falke, Leonardo~FR Ribeiro, Prasetya~Ajie Utama, Ido Dagan, and Iryna
  Gurevych. 2019.
\newblock {Ranking generated summaries by correctness: An interesting but
  challenging application for natural language inference}.
\newblock In \emph{Proceedings of the 57th Annual Meeting of the Association
  for Computational Linguistics}, pages 2214--2220.

\bibitem[{Ganchev et~al.(2010)Ganchev, Gra{\c{c}}a, Gillenwater, and
  Taskar}]{ganchev2010posterior}
Kuzman Ganchev, Joao Gra{\c{c}}a, Jennifer Gillenwater, and Ben Taskar. 2010.
\newblock Posterior regularization for structured latent variable models.
\newblock \emph{The Journal of Machine Learning Research}, 11:2001--2049.

\bibitem[{Gillick and Liu(2010)}]{gillick-liu-2010-non}
Dan Gillick and Yang Liu. 2010.
\newblock \href {https://www.aclweb.org/anthology/W10-0722} {Non-expert
  evaluation of summarization systems is risky}.
\newblock In \emph{Proceedings of the {NAACL} {HLT} 2010 Workshop on Creating
  Speech and Language Data with {A}mazon{'}s Mechanical Turk}, pages 148--151,
  Los Angeles. Association for Computational Linguistics.

\bibitem[{Goyal and Durrett(2020{\natexlab{a}})}]{goyal2020evaluating}
Tanya Goyal and Greg Durrett. 2020{\natexlab{a}}.
\newblock \href {https://www.aclweb.org/anthology/2020.findings-emnlp.322.pdf}
  {Evaluating factuality in generation with dependency-level entailment}.
\newblock In \emph{Proceedings of the 2020 Conference on Empirical Methods in
  Natural Language Processing: Findings}, pages 3592--3603.

\bibitem[{Goyal and Durrett(2020{\natexlab{b}})}]{goyal-durrett-2020-neural}
Tanya Goyal and Greg Durrett. 2020{\natexlab{b}}.
\newblock \href {https://doi.org/10.18653/v1/2020.acl-main.22} {Neural
  syntactic preordering for controlled paraphrase generation}.
\newblock In \emph{Proceedings of the 58th Annual Meeting of the Association
  for Computational Linguistics}, pages 238--252, Online. Association for
  Computational Linguistics.

\bibitem[{Hermann et~al.(2015)Hermann, Kocisky, Grefenstette, Espeholt, Kay,
  Suleyman, and Blunsom}]{hermann2015teaching}
Karl~Moritz Hermann, Tomas Kocisky, Edward Grefenstette, Lasse Espeholt, Will
  Kay, Mustafa Suleyman, and Phil Blunsom. 2015.
\newblock Teaching machines to read and comprehend.
\newblock In \emph{Advances in neural information processing systems}, pages
  1693--1701.

\bibitem[{Kang and Hashimoto(2020)}]{kang-hashimoto-2020-improved}
Daniel Kang and Tatsunori Hashimoto. 2020.
\newblock \href {https://doi.org/10.18653/v1/2020.acl-main.66} {Improved
  natural language generation via loss truncation}.
\newblock In \emph{Proceedings of the 58th Annual Meeting of the Association
  for Computational Linguistics}, pages 718--731, Online. Association for
  Computational Linguistics.

\bibitem[{Knight and Marcu(2000)}]{knight-marcu-2000-sentence-compression}
Kevin Knight and Daniel Marcu. 2000.
\newblock {Statistics-Based Summarization---Step One: Sentence Compression}.
\newblock In \emph{Proceedings of the National Conference on Artificial
  Intelligence (AAAI) and Conference on Innovative Applications of Artificial
  Intelligence (IAAI)}, pages 703--710.

\bibitem[{Kryscinski et~al.(2020)Kryscinski, McCann, Xiong, and
  Socher}]{kryscinski2020evaluating}
Wojciech Kryscinski, Bryan McCann, Caiming Xiong, and Richard Socher. 2020.
\newblock Evaluating the factual consistency of abstractive text summarization.
\newblock In \emph{Proceedings of the 2020 Conference on Empirical Methods in
  Natural Language Processing (EMNLP)}, pages 9332--9346.

\bibitem[{Lebanoff et~al.(2019)Lebanoff, Muchovej, Dernoncourt, Kim, Kim,
  Chang, and Liu}]{lebanoff-etal-2019-analyzing}
Logan Lebanoff, John Muchovej, Franck Dernoncourt, Doo~Soon Kim, Seokhwan Kim,
  Walter Chang, and Fei Liu. 2019.
\newblock \href {https://doi.org/10.18653/v1/D19-5413} {Analyzing sentence
  fusion in abstractive summarization}.
\newblock In \emph{Proceedings of the 2nd Workshop on New Frontiers in
  Summarization}, Hong Kong, China. Association for Computational Linguistics.

\bibitem[{Lewis et~al.(2020)Lewis, Liu, Goyal, Ghazvininejad, Mohamed, Levy,
  Stoyanov, and Zettlemoyer}]{lewis-etal-2020-bart}
Mike Lewis, Yinhan Liu, Naman Goyal, Marjan Ghazvininejad, Abdelrahman Mohamed,
  Omer Levy, Veselin Stoyanov, and Luke Zettlemoyer. 2020.
\newblock \href {https://doi.org/10.18653/v1/2020.acl-main.703} {{BART}:
  Denoising sequence-to-sequence pre-training for natural language generation,
  translation, and comprehension}.
\newblock In \emph{Proceedings of the 58th Annual Meeting of the Association
  for Computational Linguistics}, Online. Association for Computational
  Linguistics.

\bibitem[{Lux et~al.(2020)Lux, Sappelli, and Larson}]{lux-etal-2020-truth}
Klaus-Michael Lux, Maya Sappelli, and Martha Larson. 2020.
\newblock \href {https://www.aclweb.org/anthology/2020.eval4nlp-1.1} {{Truth or
  Error? Towards systematic analysis of factual errors in abstractive
  summaries}}.
\newblock In \emph{Proceedings of the First Workshop on Evaluation and
  Comparison of NLP Systems}, pages 1--10, Online. Association for
  Computational Linguistics.

\bibitem[{Mao et~al.(2020)Mao, Ren, Ji, and Han}]{mao2020constrained}
Yuning Mao, Xiang Ren, Heng Ji, and Jiawei Han. 2020.
\newblock {Constrained Abstractive Summarization: Preserving Factual
  Consistency with Constrained Generation}.
\newblock In \emph{arXiv preprint 2010.12723}.

\bibitem[{Maynez et~al.(2020)Maynez, Narayan, Bohnet, and
  Mcdonald}]{maynez2020}
Joshua Maynez, Shashi Narayan, Bernd Bohnet, and Ryan~Thomas Mcdonald. 2020.
\newblock On faithfulness and factuality in abstractive summarization.
\newblock In \emph{Proceedings of The 58th Annual Meeting of the Association
  for Computational Linguistics (ACL)}.

\bibitem[{Nallapati et~al.(2016)Nallapati, Zhou, dos Santos, Gulcehre, and
  Xiang}]{nallapati2016abstractive}
Ramesh Nallapati, Bowen Zhou, Cicero dos Santos, Caglar Gulcehre, and Bing
  Xiang. 2016.
\newblock Abstractive text summarization using sequence-to-sequence rnns and
  beyond.
\newblock In \emph{Proceedings of The 20th SIGNLL Conference on Computational
  Natural Language Learning}, pages 280--290.

\bibitem[{Narayan et~al.(2018)Narayan, Cohen, and Lapata}]{narayan2018don}
Shashi Narayan, Shay~B Cohen, and Mirella Lapata. 2018.
\newblock Don’t give me the details, just the summary! topic-aware
  convolutional neural networks for extreme summarization.
\newblock In \emph{Proceedings of the 2018 Conference on Empirical Methods in
  Natural Language Processing}, pages 1797--1807.

\bibitem[{Nenkova and Passonneau(2004)}]{nenkova-passonneau-2004-evaluating}
Ani Nenkova and Rebecca Passonneau. 2004.
\newblock \href {https://www.aclweb.org/anthology/N04-1019} {Evaluating content
  selection in summarization: The pyramid method}.
\newblock In \emph{Proceedings of the Human Language Technology Conference of
  the North {A}merican Chapter of the Association for Computational
  Linguistics: {HLT}-{NAACL} 2004}, pages 145--152, Boston, Massachusetts, USA.
  Association for Computational Linguistics.

\bibitem[{Nie et~al.(2019)Nie, Chen, and Bansal}]{nie2019combining}
Yixin Nie, Haonan Chen, and Mohit Bansal. 2019.
\newblock Combining fact extraction and verification with neural semantic
  matching networks.
\newblock In \emph{Proceedings of the AAAI Conference on Artificial
  Intelligence}, volume~33, pages 6859--6866.

\bibitem[{Song et~al.(2020{\natexlab{a}})Song, Lebanoff, Guo, Qiu, Xue, Li, Yu,
  and Liu}]{song2020joint}
Kaiqiang Song, Logan Lebanoff, Qipeng Guo, Xipeng Qiu, Xiangyang Xue, Chen Li,
  Dong Yu, and Fei Liu. 2020{\natexlab{a}}.
\newblock Joint parsing and generation for abstractive summarization.
\newblock In \emph{Proceedings of the AAAI Conference on Artificial
  Intelligence}, volume~34, pages 8894--8901.

\bibitem[{Song et~al.(2020{\natexlab{b}})Song, Wang, Feng, Liu, and
  Liu}]{song2020controlling}
Kaiqiang Song, Bingqing Wang, Zhe Feng, Ren Liu, and Fei Liu.
  2020{\natexlab{b}}.
\newblock Controlling the amount of verbatim copying in abstractive
  summarization.
\newblock In \emph{Proceedings of the AAAI Conference on Artificial
  Intelligence}, volume~34, pages 8902--8909.

\bibitem[{Thorne et~al.(2018)Thorne, Vlachos, Christodoulopoulos, and
  Mittal}]{thorne-etal-2018-fever}
James Thorne, Andreas Vlachos, Christos Christodoulopoulos, and Arpit Mittal.
  2018.
\newblock \href {https://doi.org/10.18653/v1/N18-1074} {{FEVER}: a large-scale
  dataset for fact extraction and {VER}ification}.
\newblock In \emph{Proceedings of the 2018 Conference of the North {A}merican
  Chapter of the Association for Computational Linguistics: Human Language
  Technologies, Volume 1 (Long Papers)}, New Orleans, Louisiana. Association
  for Computational Linguistics.

\bibitem[{Tian et~al.(2019)Tian, Narayan, Sellam, and
  Parikh}]{tian2019sticking}
Ran Tian, Shashi Narayan, Thibault Sellam, and Ankur~P Parikh. 2019.
\newblock Sticking to the facts: Confident decoding for faithful data-to-text
  generation.
\newblock \emph{arXiv preprint arXiv:1910.08684}.

\bibitem[{Wang et~al.(2020)Wang, Cho, and Lewis}]{wang2020asking}
Alex Wang, Kyunghyun Cho, and Mike Lewis. 2020.
\newblock {Asking and Answering Questions to Evaluate the Factual Consistency
  of Summaries}.
\newblock In \emph{Proceedings of the 58th Annual Meeting of the Association
  for Computational Linguistics}.

\bibitem[{Wolf et~al.(2019)Wolf, Debut, Sanh, Chaumond, Delangue, Moi, Cistac,
  Rault, Louf, Funtowicz, and Brew}]{Wolf2019HuggingFacesTS}
Thomas Wolf, Lysandre Debut, Victor Sanh, Julien Chaumond, Clement Delangue,
  Anthony Moi, Pierric Cistac, Tim Rault, R'emi Louf, Morgan Funtowicz, and
  Jamie Brew. 2019.
\newblock Huggingface's transformers: State-of-the-art natural language
  processing.
\newblock \emph{ArXiv}, abs/1910.03771.

\bibitem[{Woodsend and Lapata(2012)}]{woodsend-lapata-2012-multi-aspect}
Kristian Woodsend and Mirella Lapata. 2012.
\newblock {Multiple Aspect Summarization Using Integer Linear Programming}.
\newblock In \emph{Proceedings of the Joint Conference on Empirical Methods in
  Natural Language Processing (EMNLP) and Computational Natural Language
  Learning (CoNLL)}.

\bibitem[{Zhang et~al.(2018)Zhang, Yao, and Yan}]{zhang2018abstractiveness}
Fang-Fang Zhang, Jin-ge Yao, and Rui Yan. 2018.
\newblock On the abstractiveness of neural document summarization.
\newblock In \emph{Proceedings of the 2018 Conference on Empirical Methods in
  Natural Language Processing}, pages 785--790.

\bibitem[{Zhang et~al.(2020{\natexlab{a}})Zhang, Zhao, Saleh, and
  Liu}]{pegasus2020}
Jingqing Zhang, Yao Zhao, Mohammad Saleh, and Peter Liu. 2020{\natexlab{a}}.
\newblock {PEGASUS}: Pre-training with extracted gap-sentences for abstractive
  summarization.
\newblock In \emph{Proceedings of the International Conference on Machine
  Learning}.

\bibitem[{Zhang et~al.(2020{\natexlab{b}})Zhang, Merck, Tsai, Manning, and
  Langlotz}]{zhang-etal-2020-optimizing}
Yuhao Zhang, Derek Merck, Emily Tsai, Christopher~D. Manning, and Curtis
  Langlotz. 2020{\natexlab{b}}.
\newblock \href {https://doi.org/10.18653/v1/2020.acl-main.458} {Optimizing the
  factual correctness of a summary: A study of summarizing radiology reports}.
\newblock In \emph{Proceedings of the 58th Annual Meeting of the Association
  for Computational Linguistics}, Online. Association for Computational
  Linguistics.

\bibitem[{Zhao et~al.(2020)Zhao, Cohen, and Webber}]{zhao2020reducing}
Zheng Zhao, Shay~B Cohen, and Bonnie Webber. 2020.
\newblock Reducing quantity hallucinations in abstractive summarization.
\newblock In \emph{Findings of EMNLP}.

\end{thebibliography}
\bibliographystyle{acl_natbib}

\appendix

\section{Manual Annotation of Errors}
\label{app:annotation}
In Section \ref{sec:error-discussion}, we outline the error distributions of multiple factuality datasets. The distributions were obtained by combing manual annotations from two authors of this paper. On a common set of 50 summaries annotated by both authors, we observe the following: (1) Both authors agreed on what spans/hallucinations within a summary constitute an error 74\% of the times. (2) In cases when both authors marked a common span as erroneous, they agreed on the error category 84\% of the time. 

\section{Synthetic Dataset Performance on held-out samples}
\label{app:cnndm-held-out}
Section \ref{sec:exp-synthetic-datasets} evaluates the performance of models trained on the synthetic datasets on human annotated test sets for two summarization domains. Here, we report the model performance on held-out tests datasets that are constructed in the same way as the training datasets. Table \ref{table:cnn-synthetic} presents these results. For both domains, we see that the models report very high performance indicating that they are able to fit the distribution of the synthetic domain. However, we see in Section 5.1 that the performance is significantly lower on actual generation outputs, with close to majority label baseline performance on the more challenging \textsc{XSum} domain. This means that the two datasets have inherently different error distributions. 

\begin{table}[h]
\small
\centering
\begin{tabular}{r|cc}
\toprule
 & Ent-C & Gen-C \\   \midrule
\multicolumn{3}{l}{\textsc{Cnn/Dm}} \\ \midrule
Sent-Factuality & 96.4  & 91.2  \\
DAE & 95.4  & 97.3  \\
DAE-Weak & 94.8  & 97.8 \\ \midrule
\multicolumn{3}{l}{\textsc{XSum}} \\ \midrule
Sent-Factuality & 96.1 & 97.9  \\
DAE & 94.3 & 97.1 \\
DAE-Weak & 95.3 & 95.9 \\
\bottomrule
\end{tabular}
\caption{Performance of factuality models when trained on synthetic factuality training datasets on their held-out test sets.}
\label{table:cnn-synthetic}
\end{table}

\begin{figure}[h]
\centering
    \includegraphics[trim=60mm 50mm 40mm 30mm,scale=0.14,clip]{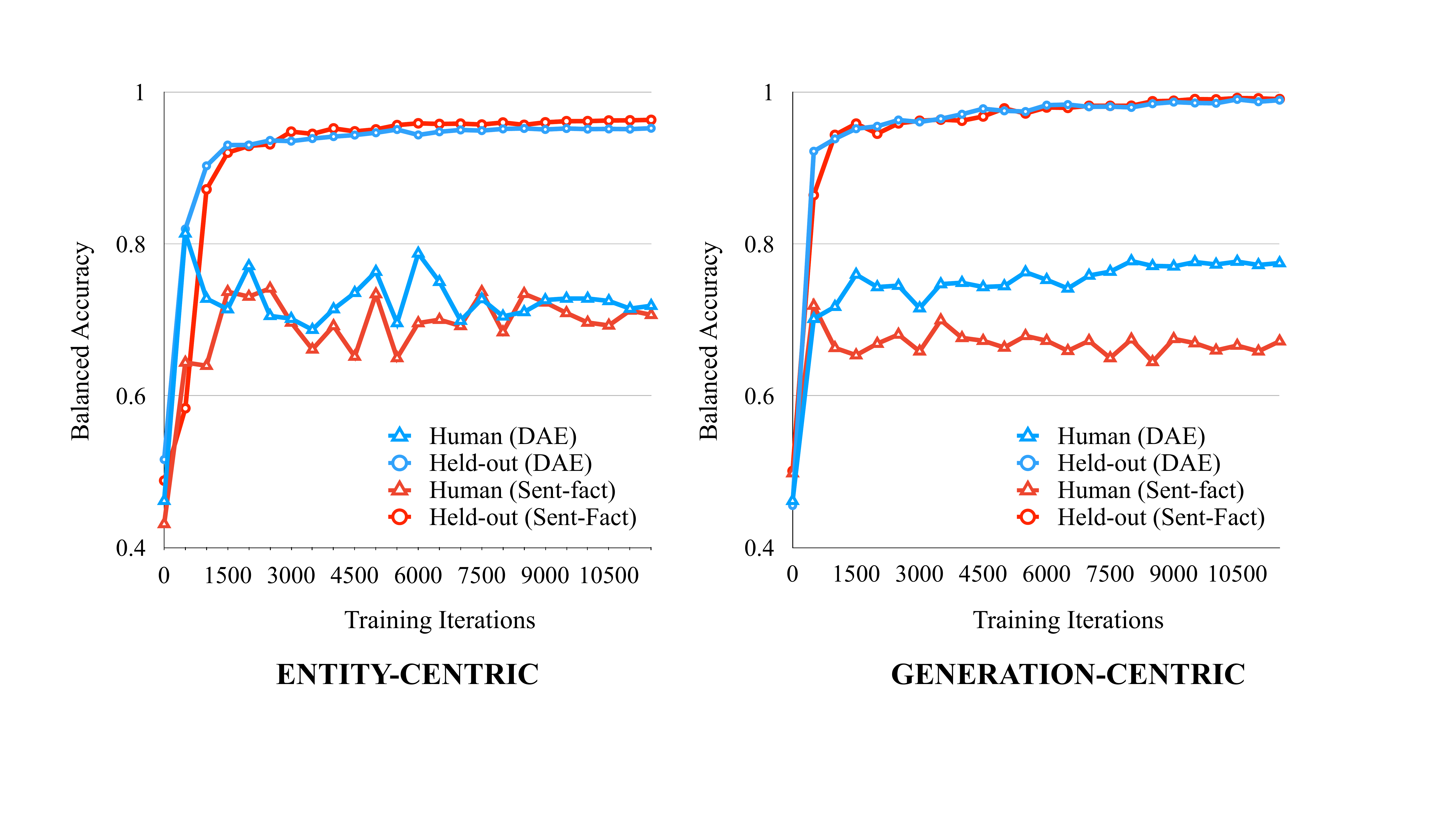}
    \caption{Performance of different train checkpoints on a held-out dataset and on the human annotated dev set for models trained on the synthetic data in the \textsc{Cnn/Dm} domain.}
    \label{fig:train-checkpoint-graphs}
\end{figure}

Figure \ref{fig:train-checkpoint-graphs} shows the balanced accuracy values reported by the model at different points during its training, on both the synthetic and human-annotated test sets. The graph clearly shows that performance on the human annotated dataset (\textsc{Cnn/Dm}) has high variance, compared to the held-out dataset accuracies which has a steadily increasing performance.  This behavior was observed for both ENT-C and GEN-C domains; however, ENT-C exhibited more variance. This indicates that the synthetic datasets are targeting a different error distribution, and optimizing for the synthetic distribution does not necessarily improve results on the actual generation errors. 

\section{Transferability of human annotations across generation models within the same domain}
\label{app:xsum-transferability}
In Section 5.2, we demonstrate that for highly abstractive domains like \textsc{XSum}, we require human annotated data to train factuality models. However, even within the same summarization domain (say \textsc{XSum}), it is prohibitively expensive to collect human annotations for each summarization model that we may wish to evaluate. Here, we investigate whether the factuality annotations collected for one summarization model be used to identify factuality errors in summaries generated by other models. These experiments are done within the same domain (\textsc{XSum})

We create new training and test sets from the \textsc{XSum-Human} dataset. We create 2 types of training datasets for each of the 5 models annotated in that dataset: (1) \textbf{All-models} train set: This contains (A,S) pairs from all models, including the models being evaluated (2000 pairs from other models, 200 pairs from same model) and (2) \textbf{Other-models} train set: This contains (A, S) pairs from the rest of the models (2000 pairs). Evaluation is done on the remaining 300 (A,S) pairs for each summarization model. We train the best performing factuality model, i.e. the DAE model for all these settings.  

\begin{table}[H]
\renewcommand{\tabcolsep}{5pt}
\small
\centering
\begin{tabular}{r|cccc}
\toprule
 & \textsc{BertS2S} & \textsc{PtGen} & \textsc{TConvS2S} & \textsc{TranS2S} \\ \midrule
All & 79.6 & 75.8 & 76.7 & 84.5  \\
Others & 82.3 & 77.0 & 74.1  & 85.3\\
\bottomrule
\end{tabular}
\caption{Performance of models trained on All-models dataset vs Other-models dataset.} 
\label{table:model-transferrability}
\end{table}

Results are outlined in table \ref{table:model-transferrability}. These show that the performance is similar for both All-models and Other-models settings for all models considered. This indicates that for the given set of summarization models considered (all trained on the same summarization training dataset), human annotations from one generation model can be used to evaluate factuality for other models. 

\section{Implementation Details}
All our factuality models are trained by fine-tuning the pre-trained \textsc{Electra} (electra-base-discriminator,  110M parameters) model. We perform 5 hyper parameter trials to select the best set of hyper parameters, varying the learning rate. The final hyper-parameters are: 

\begin{table}[h]
    \small
    \begin{tabular}{l|l}\toprule
        Implementation Library & transformers \cite{Wolf2019HuggingFacesTS} \\
        Computing Infrastructure & 32GB NVIDIA V100 GPU \\
        Max Seq Length & 512 \\
        Optimizer & Adam\\
        Optimizer Params & $\beta=(0.9, 0.999), \epsilon=10^{-8}$ \\
        Learning Rate Decay & Linear \\
        Learning rate & 2e-5\\
        Weight Decay & 0 \\
        Warmup Steps & 0 \\
        Maximum Gradient Norm & 1 \\
        Batch size & 8 \\
        Epochs & 3 \\
         \bottomrule
    \end{tabular}
    \caption{Hyperparameters used for fine-tuning the factuality models. }
    \label{table:params-fact}
\end{table}
For models with high variance (sent-factuality model from section 5.2), we report average of 3 runs by initializing with a random seed.

The hyperparameters for training the \textsc{Bart} summarization models are given in Table \ref{table:params-bart}. Parameter settings used  during decoding to generate summaries on the test set are also included
\begin{table}[h]
    \small
    \begin{tabular}{l|l}\toprule
        \multicolumn{2}{l}{For training} \\ \midrule
        Implementation Library & transformers \cite{Wolf2019HuggingFacesTS} \\
        Computing Infrastructure & 32GB NVIDIA V100 GPU \\
        Max Input Seq Length & 512 \\
        Max Output Seq Length & 128 \\
        Optimizer & Adam\\
        Optimizer Params & $\beta=(0.9, 0.999), \epsilon=10^{-8}$ \\
        Learning Rate Decay & Linear \\
        Learning rate & 2e-5\\
        Weight Decay & 0 \\
        Warmup Steps & 0 \\
        Maximum Gradient Norm & 1 \\
        Batch size & 8 \\
        Epochs & 10 \\ \midrule
        \multicolumn{2}{l}{For decoding} \\ \midrule
        Num beams & 6 \\
        Length Penalty & 2 \\
        No repetition size & 3-grams \\
        Min-Length & 10 \\
        Max Length & 60 \\ \bottomrule
    \end{tabular}
    \caption{Hyperparameters used for fine-tuning and decoding using the \textsc{Bart}-based summarization models.}
    \label{table:params-bart}
\end{table}

\section{Human Study}
\label{app:mtruk}
\begin{figure*}[h]
\centering
\includegraphics[trim=140mm 50mm 100mm 30mm,scale=0.47,clip]{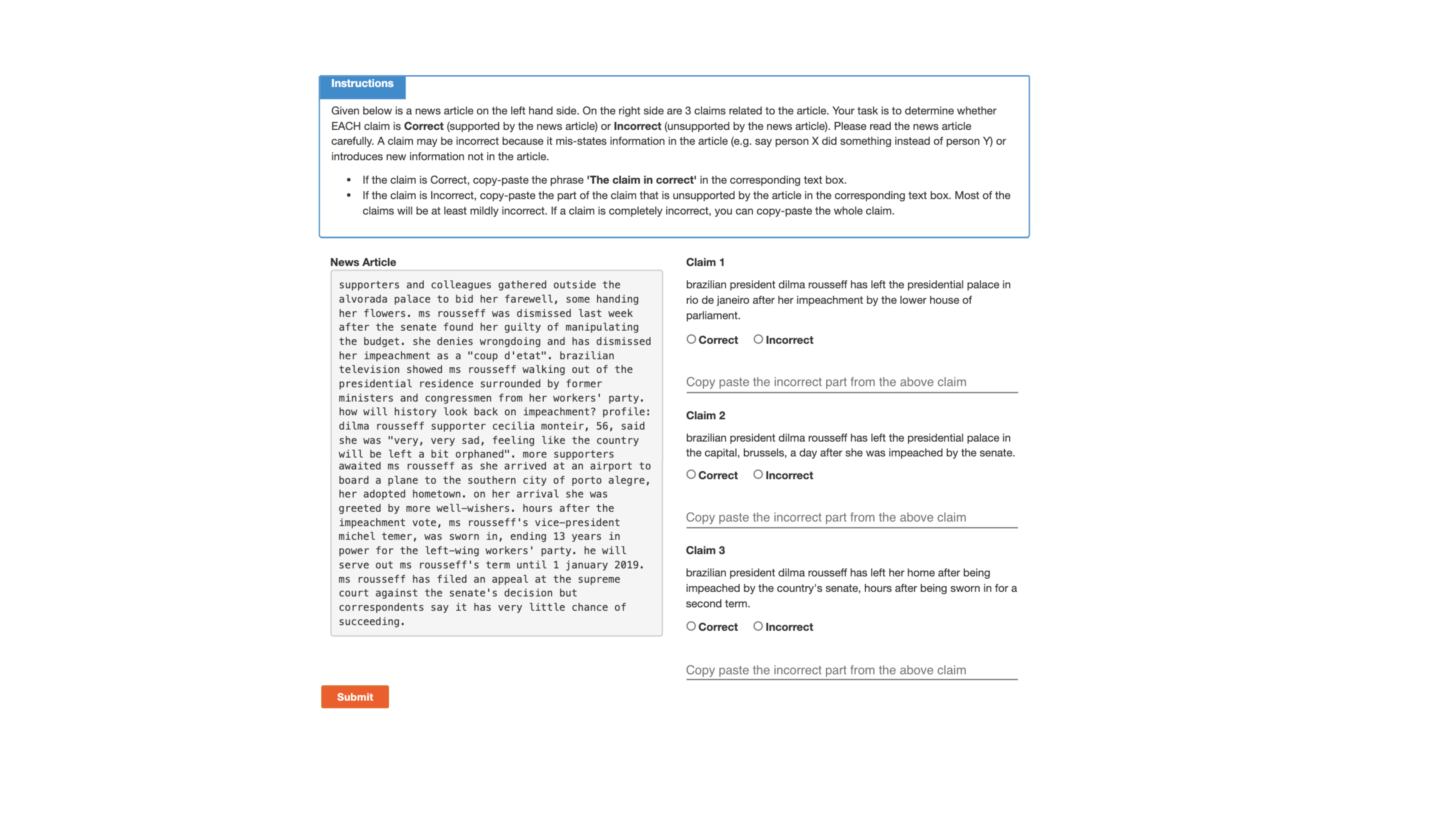}
\caption{Screenshot of the Mechanical Turk experiment. Given an input articles, the annotators were tasked with evaluating the factuality of 3 model generated summaries on a binary scale.}\vspace*{5in}
\label{fig:mturk}
\end{figure*}

Figure \ref{fig:mturk} provides an screenshot of the Amazon Mechanical Turk tasks used to obtain human judgements for factuality of generated summaries, as outlined in Section \ref{sec:downstream}. Workers were presented with a source article and $3$ corresponding summaries. Each of these summaries were marked as Factual or Non-Factual. Additionally, they were asked to highlight the span within the summary that was erroneous.  

\end{document}